\documentclass[10pt,twocolumn,letterpaper]{article}

\usepackage{cvpr}              

\usepackage{graphicx}
\usepackage{amsmath}
\usepackage{amssymb}
\usepackage{booktabs}
\usepackage{enumitem}
\usepackage{svg}
\usepackage{mathtools}

\usepackage[pagebackref,breaklinks,colorlinks]{hyperref}

\usepackage[capitalize]{cleveref}
\crefname{section}{Sec.}{Secs.}
\Crefname{section}{Section}{Sections}
\Crefname{table}{Table}{Tables}
\crefname{table}{Tab.}{Tabs.}

\def\cvprPaperID{} 
\def\confName{CVPR}
\def\confYear{2022}

\begin{document}

\title{Web Neural Network with Complete DiGraphs}

\author{Frank Li\\
University of Washington\\
Seattle, WA, United States\\
{\tt\small angli23@cs.washington.edu}}
\maketitle

\begin{abstract}
   This paper introduces a new neural network model that aims to mimic the biological brain more closely by structuring the network as a complete directed graph that processes continuous data for each timestep. Current neural networks have structures that vaguely mimic the brain structure, such as neurons, convolutions, and recurrence. The model proposed in this paper adds additional structural properties by introducing cycles into the neuron connections and removing the sequential nature commonly seen in other network layers. Furthermore, the model has continuous input and output, inspired by spiking neural networks, which allows the network to learn a process of classification, rather than simply returning the final result.
\end{abstract}

\section{Introduction}
\label{sec:intro}

Today, there exists numerous neural network structures such as fully-connected, convolutional, and graph neural networks \cite{zhou_graph_2021}. But none of them fully mimic the biological neural network structure; graph neural networks are probably the most similar in terms of visual structure. Previously, most neural network structures followed a linear process, like multi-layer perceptron; consequently, researchers have come up with recursive solutions such as recurrent networks. However, that still does not capture the complexity of a biological brain.
Firstly, in reality, inputs for neurons are not scalar, they are instead wave signals with synapses, which are heavily emphasized in the field of spiking neural networks where consecutive activation has different behavior from singular activation \cite{chowdhury_one_2021}. Moreover, connections of neurons in a biological neural network are also not necessarily linear or recursively patterned. They are instead shaped as a dense, highly interconnected graph with unpredictable connections triggered by neuron activation. Therefore, the neural network model proposed in this paper will try to mimic such  structure by not supplying the network with any predefined wiring.

\subsection{Idea}

This project seeks to propose a neural network model that leverages an open sea of neurons with no predefined linear structure, like a web. The web of neurons will begin as a complete directed graph, where each neuron can connect to every other neuron, even cyclically and recursively. This idea is heavily inspired by randomly wired neural networks  \cite{xie_exploring_2019}. Except, instead of generating random graphs, the model in this paper will generate complete directed graphs using hyperparameters (e.g., number of neurons) and aim to handle continuous data. By allowing cyclic and recursive connections, the model can possess contexts of previous timesteps, because the intermediate results of the previous timesteps remain in the web of neurons, and as time passes, data from older timesteps will naturally be overwritten by newer data, hence simulating a decay of knowledge contexts. Therefore, this model should be functionally similar to LSTMs by preserving previous contexts while specializing in continuous data like spiking neural networks.

\subsection{Possible Challenges}

The most difficult part of this project will likely be the technical implementation of the network since I am not aware of any frameworks that currently support a network as such. This is because the proposed neural network is not sequential, but rather a collection of freely connected neurons. For instance, in PyTorch, a common implementation of networks would be to use a \verb|nn.Sequential| component where values are propagated through the layers sequentially. However, that will not work for the model proposed in this paper since the forward propagation is not sequential. Another challenge will be computing resources. The computational complexity of the training and inference steps can be very inefficient. For $E$ epochs, $T$ timesteps, and $Q$ neurons, calculating the output for every neuron will have complexity $O(ETQ)$ with very high $Q$, which can be extremely slow. Therefore, it may be the case that the long processing time will limit the scope of the project to only be able to train on simple datasets for a small number of epochs.

\subsection{Expectations}

It can be predicted that the model will be computationally demanding due to the nature of the neural network – forward and backward passes will have more complex computational graphs. Hence, having less cost-performance efficiency. However, the performance of the network will likely be greater than that of a vanilla fully-connected neural network. Finally, there may be interesting conclusions and visualizations that can be produced from the output data. For example, there can be a situation such that, at timestep 1, the network classifies the input as digit 1, while at the final timestep, it is classified as digit 9, thereby observing a reasoning process for how the neural network’s conclusions move from one to another.

\section{Related Work}

\noindent\textbf{Randomly Wired Neural Networks} \cite{xie_exploring_2019} is a neural network model that utilizes stochastic network generation to generate random graphs as neural network layers. Xie et. al. have found the technique to be a valuable neural architecture search mechanism that can create models that are competitive with ResNet-50 and ResNeXt-50. The web neural network in this paper is heavily inspired by the random wiring of neurons in a graph-style network.\medskip

\noindent\textbf{LSTM} \cite{staudemeyer_understanding_2019} is a network model that solves the vanishing gradient problem in recurrent neural networks. By retaining knowledge of embeddings from previous tokens, LSTMs are able to generate tokens based on contextual knowledge. It is often used to model continuous data such as time series forecasting, which takes timesteps into account. \medskip

\noindent\textbf{Spiking Neural Networks} \cite{chowdhury_one_2021} is a biologically inspired model that, instead of having static constant neuron activation in traditional neural networks, stores an internal temporal state in neurons such that the timing of the synapses changes how the neuron behaves. For example, if a value is passed into a neuron it may not do anything immediately, but it stores the value in its state; then if no further value is received for a while, the value will decay, causing the neuron to return to its original state. However, if another value is passed in immediately after the first value, the neuron could reach its ``action potential'' and hence propagate to other neurons.

\section{Methodology}

The proposed model will be trained on small numerical datasets from Kaggle and the MNIST dataset to illustrate and compare the performance of the web neural network. This paper will also introduce different variations of the network such as ones preceded by a convolutional layer for classifying images.

\subsection{Model}

\noindent\textbf{Hyperparameters} of this model include: number of neurons ($Q$), number of input neurons ($I$), number of output neurons ($O$), and max timesteps ($T$). We need the max timestep as a hyperparameter because, since the network is cyclical and recursive, there does not exist a ``final layer'' that the neural network will output to. Instead, we will record the selected output neurons at each timestep, and reason about the recorded continuous output data. Intuitively, increasing the number of timesteps will be analogous to increasing the “depth” of the network, since there will be more propagations between neurons. Moreover, we can also introduce other hyperparameters to support extra training techniques such as dropout (i.e., dropout rate $R$).\medskip

\begin{figure}[h]
	\centering
	\resizebox{\linewidth}{!}{
\begingroup%
  \makeatletter%
  \providecommand\color[2][]{%
    \errmessage{(Inkscape) Color is used for the text in Inkscape, but the package 'color.sty' is not loaded}%
    \renewcommand\color[2][]{}%
  }%
  \providecommand\transparent[1]{%
    \errmessage{(Inkscape) Transparency is used (non-zero) for the text in Inkscape, but the package 'transparent.sty' is not loaded}%
    \renewcommand\transparent[1]{}%
  }%
  \providecommand\rotatebox[2]{#2}%
  \newcommand*\fsize{\dimexpr\f@size pt\relax}%
  \newcommand*\lineheight[1]{\fontsize{\fsize}{#1\fsize}\selectfont}%
  \ifx\svgwidth\undefined%
    \setlength{\unitlength}{355.04220317bp}%
    \ifx\svgscale\undefined%
      \relax%
    \else%
      \setlength{\unitlength}{\unitlength * \real{\svgscale}}%
    \fi%
  \else%
    \setlength{\unitlength}{\svgwidth}%
  \fi%
  \global\let\svgwidth\undefined%
  \global\let\svgscale\undefined%
  \makeatother%
  \begin{picture}(1,0.29538989)%
    \lineheight{1}%
    \setlength\tabcolsep{0pt}%
    \put(0,0){\includegraphics[width=\unitlength,page=1]{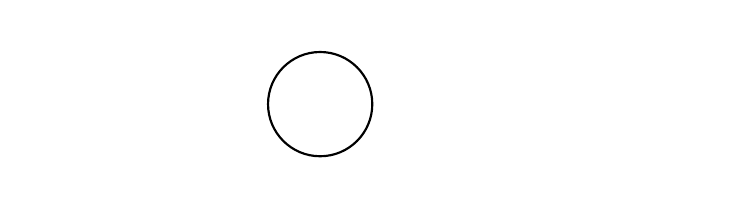}}%
    \put(0.395551,0.14528055){\color[rgb]{0,0,0}\makebox(0,0)[lt]{\lineheight{1.25}\smash{\begin{tabular}[t]{l}$F(X)$\end{tabular}}}}%
    \put(-0.00128728,0.15701828){\color[rgb]{0,0,0}\makebox(0,0)[lt]{\lineheight{1.25}\smash{\begin{tabular}[t]{l}Outputs from\\all other nodes\end{tabular}}}}%
    \put(0,0){\includegraphics[width=\unitlength,page=2]{fig_neuron.pdf}}%
    \put(0.66700836,0.16656049){\color[rgb]{0,0,0}\makebox(0,0)[lt]{\lineheight{1.25}\smash{\begin{tabular}[t]{l}Outputs to all other nodes\\on the next time step\end{tabular}}}}%
    \put(0,0){\includegraphics[width=\unitlength,page=3]{fig_neuron.pdf}}%
  \end{picture}%
\endgroup%
}
	\caption{Neuron Structure}
	\label{fig:neuron}
\end{figure}

\begin{figure}[h]
	\centering
	\resizebox{\linewidth}{!}{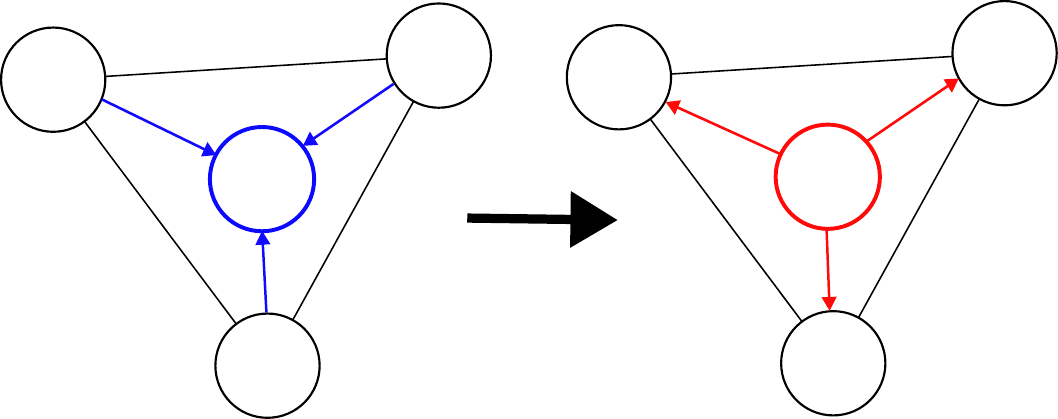}
	\caption{Neuron Activity}
	\label{fig:network}
\end{figure}

\noindent\textbf{Neurons} including input and output neurons will take continuous data, process them , and output to all other neurons at each timestep (\cref{fig:neuron}, \cref{fig:network}). They have the same connections as nodes in a complete directed graph. For non-linearity, all neuron outputs are passed into an activation function. Since input and output neurons are sampled from the entire collection of neurons, $Q\ge I+O$ must be true.\medskip

\noindent\textbf{Input and output}. The model expects continuous data, so the input data should have dimension $(N,T_{i},I,\ldots)$ where $N$ is the number of samples, $T_{i}$ is the total timestep of the data, and the remaining dimensions will be the dimensions of each input sample. Note that $T_{i}$ can be different from the $T$ in the hyperparameter.  The output neurons will also output continuous data since input loading and forward propagation will be done in timesteps. Therefore, the output will have dimension $(N,T,O,\ldots)$ where there is an output $y$ for every sample at every timestep.\medskip

\noindent\textbf{Every timestep}, each neuron will take in input from all other neurons and pass the mean of all values into $F(X)$, whose result is propagated to all other neurons. This will be repeated in all timesteps. Therefore, $F(X)$ will have input and output dimensions of $Q$. The output neuron's value will be recorded every timestep, forming a time-series data.

\subsection{Training}

To assess the model's performance on different applications, the model will be used to solve conventional numerical binary classification (titanic) and image classification (MNIST) problems. For each problem, there may also be additional network layers required to process the data, for example, convolutional layers for image processing.

\section{Implementation}

To have neurons propagate results between themselves, we can utilize an adjacency matrix of size $Q\times Q$ to represent the state of the network at time $t$. Let this state matrix be $S$. Input will have dimensions smaller than $Q$, hence, we can pad the input vector to be size $Q$ and add it to the input columns of each input neuron. Finally, output can be read from the output neurons' columns.

\subsection{Naive Solution}

With $S_t$ known, in order to calculate the state of node $i$ for time $t+1$, we can get the values for column $i$ to represent the input for node, use $F_i(X)$ to calculate the output (which also length $Q$), apply non-linearity, and finally store the results in row $i$. Repeating this for every node $i\in [0,Q]$ will create a state matrix $S_{t+1}$ where the outputs of $t$ are stored in every row, so that column $i$ in $S_{t+1}$ will be the input for node $i$ when calculating $S_{t+2}$. This way, treating the calculation of one node as a unit of operation, forward propagation for this model will have time complexity $O(TQ)$. This can cause the training time of this model to be extremely slow since the complexity with the number of epochs $E$ will be $O(ETQ)$ with a large $Q$ since $Q\ge I+O$. Not only that, in PyTorch, this will require $Q$ instances of \verb|nn.Linear| to be stored in the module.

\subsection{Vectorized Solution}

To make the training faster, we can vectorize the state calculation so we don't have to iterate through the calculation for every node. This can be done by utilizing the batch multiplication function -- \verb|torch.matmul| in PyTorch. To be specific, suppose that weights $W$ in $F_i(x)$ have dimension $(Q,Q)$ and the $i$-th column has dimension $Q$, then we have output with dimension $Q$. With the naive iterative solution, this calculation needed to be done $Q$ times. But with \verb|torch.matmul|, we can calculate this in batch by forming operation $(N,Q,Q,Q)$\verb|.matmul|$(N,Q,Q,1)$ which has output (batched state matrix) with dimension $(N,Q,Q,1)$ \cite{NEURIPS2019_9015}. We just need to make sure to transpose the result so that the output in $t$ is treated as input in the calculation for $t+1$. This reduces the training complexity to $O(ET)$, which showed a significant difference in training time compared to the naive solution when training on the MNIST dataset which used $500$ nodes.

\section{Experiments}

\subsection{Titanic Dataset}

The Titanic dataset from Kaggle is a binary classification problem where the data includes passenger information such as sex, age, cabin, etc. and the goal is to predict if the passenger will survive (true or false). The goal of training a model using the Titanic dataset is to verify that the model and implementation are functional in practice. Before training the model, we first pre-processed the input data to embed categorical text columns into numerical data. We also added a sigmoid function to the output to scale it to be probabilities within $[0,1]$, where prediction of $0.5$ or above were treated as true and otherwise false. We also chose leaky ReLU to be the activation function to avoid the dead ReLU problem. Finally, for training, we used mini-batch stochastic gradient descent with AdamW optimizer and an exponential learning rate scheduler. The hyperparameters were:

\begin{itemize}
    \item Training: $\text{epochs}=40$, $\text{batch size}=64$
    \item Web Layer: $Q=15$, $I=8$, $O=1$, $T=30$
    \item AdamW Optimizer: $\gamma=0.01$, $\lambda=0.001$
    \item Learning Rate Scheduler: $\gamma=0.9$
\end{itemize}

\begin{figure}[h]
	\centering
    \includegraphics[]{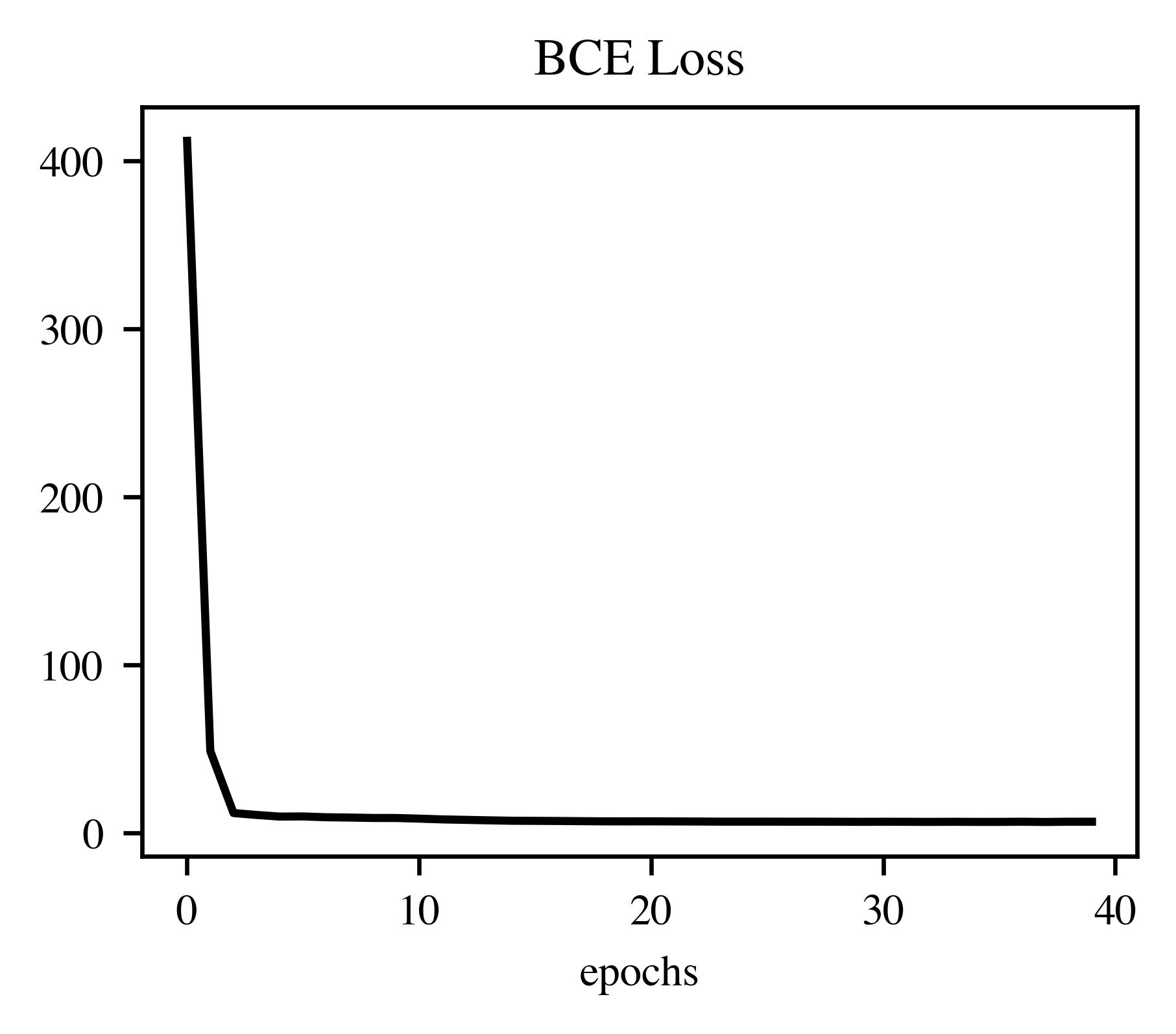}
	\caption{Titanic Training Loss}
	\label{fig:bceloss}
\end{figure}

\begin{figure}[h]
	\centering
    \includegraphics[]{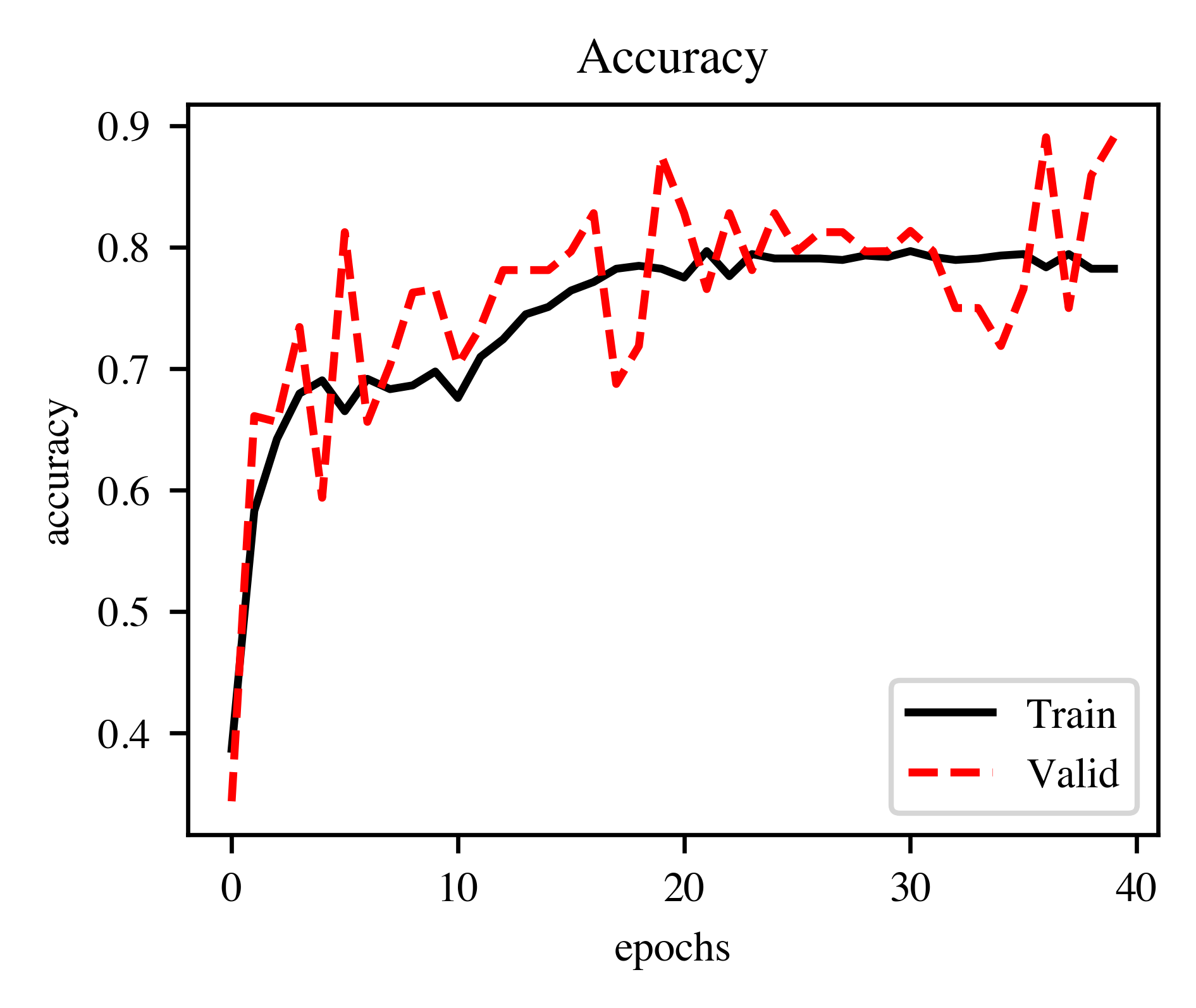}
	\caption{Titanic Accuracy}
	\label{fig:titanic}
\end{figure}

As the result \cref{fig:bceloss} and \cref{fig:titanic} above illustrate, we have clear trends of decreasing loss and increasing training and validation accuracies over time. Therefore, the model is indeed functional and able to classify data through supervised learning. The above neural network was able to reach a validation accuracy of 89\% at the end of training, and upon submitting the predictions for the test inputs to Kaggle, it returned a test accuracy of 76\%.

\subsection{MNIST Dataset}

For the MNIST dataset, before the web layer, we first used three convolution layers, each followed by leaky ReLU activation, to extract features from the images. The input was directly passed through the convolutional layers, and then the output of the third convolution layer was flattened and passed into the web layer. For the final output, a softmax function was applied to the ten values, and the digit with the highest probability value was treated as the prediction. Again, mini-batch SGD with AdamW optimizer was used for training. The hyperparameters were:

\begin{itemize}
    \item Training: $\text{epochs}=5$, $\text{batch size}=128$
    \item Conv Layer 1: $C_{in}=1$, $C_{out}=16$, $K=3$
    \item Conv Layer 2: $C_{in}=16$, $C_{out}=4$, $K=3$ 
    \item Conv Layer 3: $C_{in}=4$, $C_{out}=1$, $K=3$ 
    \item Web Layer: $Q=500$, $I=441$, $O=10$, $T=5$
    \item AdamW Optimizer: $\gamma=0.001$
\end{itemize}

\begin{figure}[h]
	\centering
    \includegraphics[]{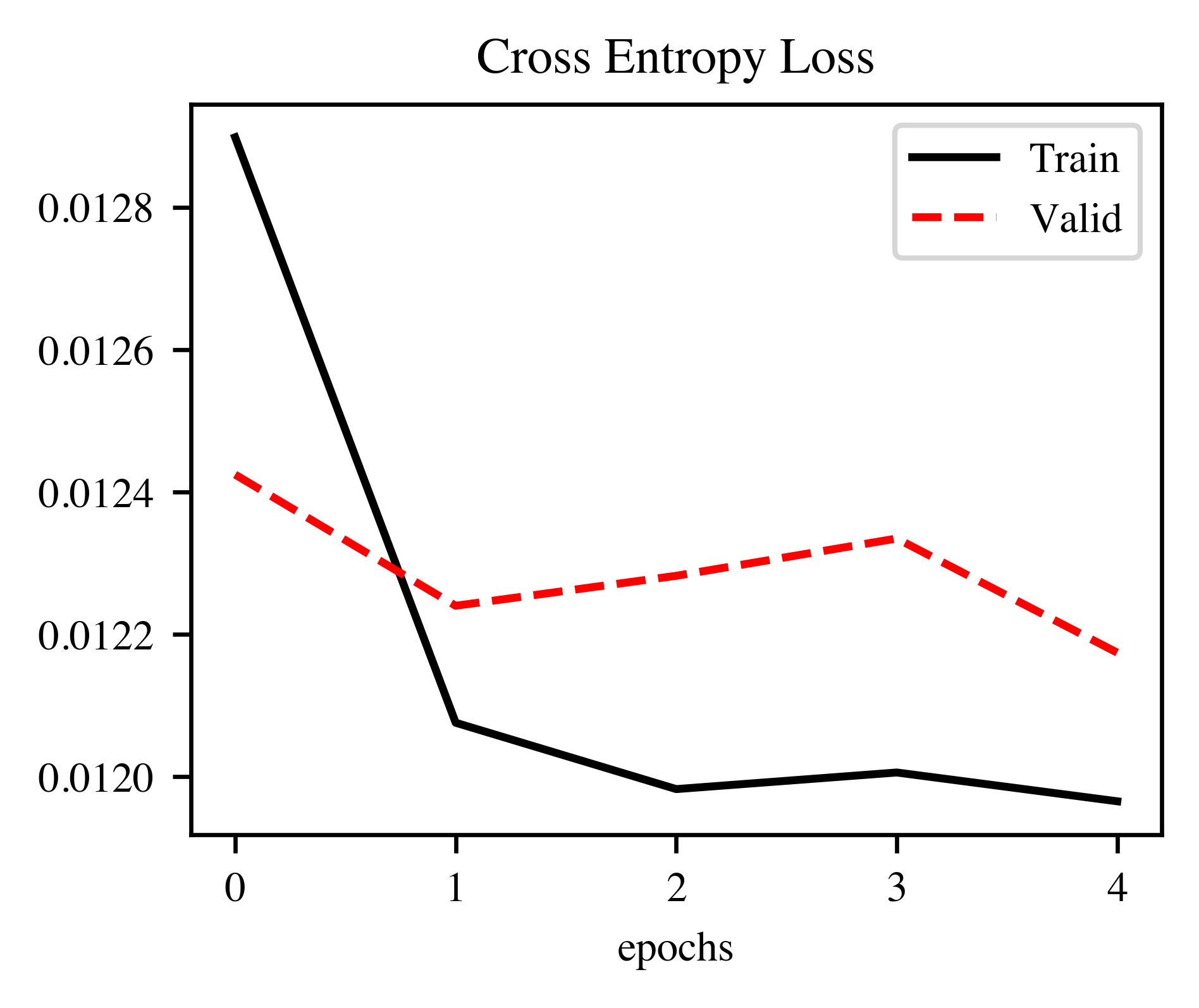}
	\caption{MNIST Training Loss}
	\label{fig:celoss}
\end{figure}

\begin{figure}[h]
	\centering
    \includegraphics[]{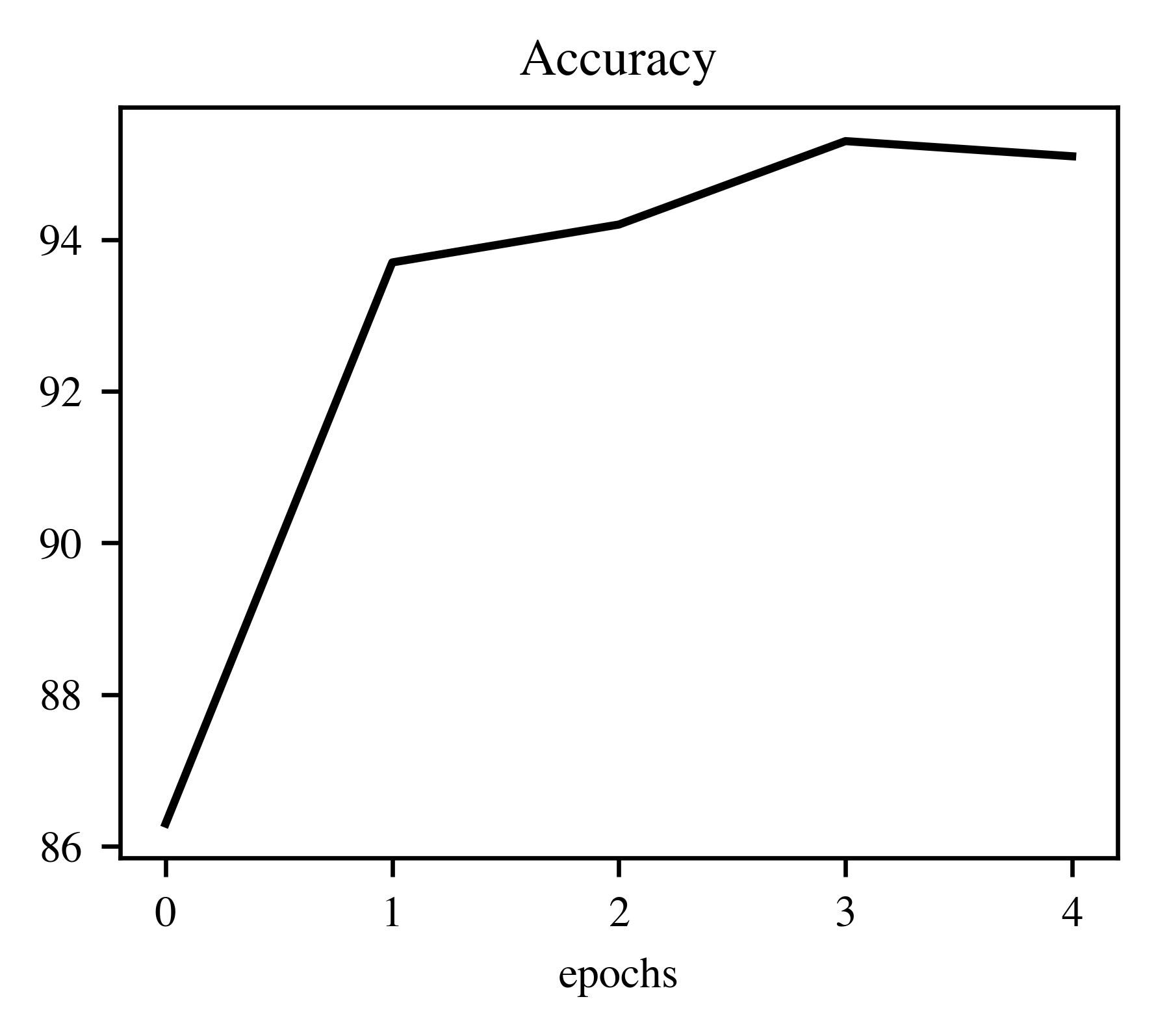}
	\caption{MNIST Accuracy}
	\label{fig:mnist}
\end{figure}

The model was also functional in this problem; over five epochs, the model was able to reach a validation accuracy of 95\% and test accuracy of 94\%. We can also see that the starting accuracy at the end of the first epoch was already very high at above 86\% and the accuracy quickly plateaus at around 95\%. This may be due to the fact that there were three convolutional layers before the web network which is known to be very effective in classifying MNIST images; however, since the final classification layer is the web layer, it must have contributed to the high accuracy as well.

\begin{figure}[h]
	\centering
    \includegraphics[width=\linewidth]{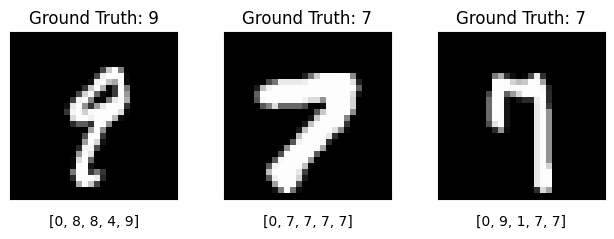}
	\caption{MNIST Prediction Per Timestep}
	\label{fig:timestep}
\end{figure}

For the web layer model classifying MNIST images, we added an array to store the historical predictions made after each timestep. \cref{fig:timestep} shows three example images that the model evaluated with the images group truth shown on the top and the history array of length 5 at the bottom. There are some interesting observations to be made from those history arrays that seemingly portray the ``thinking process'' of the model and how it reaches the conclusions.

Focusing on the first image, we can see that the prediction changed from 0 to 8, 8 to 4, and 4 to 9. The predictions 0, 8, and 4 are reasonable as they all share some similar structure with 9, especially in that specific image. This change in prediction illustrates a clear process. For confusing images like this, the model predicts similar but incorrect digits in earlier timesteps and reaches the correct conclusion at the final timestep. This behavior is further proven by the second and third example images with a ground truth of 7. The first 7 has a very distinct structure that does not share any properties with other digits, while the second 7 has an extra line on the left that makes it similar to a 9 and a very straight line downwards that makes it similar to a 1. The history arrays for those two images are in agreement with this observation. For the first 7, the model predicts correctly at timestep 2 and never changes its conclusion, while for the second 7, the model predicts 9 and 1 first, then finally 7. Thus, from \cref{fig:timestep}, we see that this MNIST model has an explicit and explainable classification process.

\section{Discussion}

However, there were some limitations involving the experiments conducted in this paper mainly due to time and resource issues. The hyperparameters used in the models are not fully optimized, including those that are not for the web layers, which means that the results may not be fully representative of the performance of the web neural network. Moreover, the experiments could also benefit from comparison with existing models such as training time, accuracy, and number of parameters. However, that could be done in future papers discussing the performance of web neural networks which was not the main focus of this paper.

Furthermore, we limited the scope of experiments to only include classification problems with constant input since the goal was to prove that the model is functional and showcase some unique properties. Hence, there are still a lot of applications and properties to be explored like:\medskip

\noindent\textbf{Wiring Modifications} by changing how neurons connect. For example, we can limit the number of connections per neuron or add special cases for input and output neurons like disabling direct connection from input to output.\medskip

\noindent\textbf{Testing Effects of Hyperparameters} such as how increasing the maximum timesteps or number of nodes affects model performance in different scenarios.\medskip

\noindent\textbf{Training on Continuous Input}. Although we treated the output as continuous and interpreted it as a process, both experiments in this paper used constant input data stretched to $T$ timesteps instead of actual continuous data. Thus, more experiments can be done with time-series data such as videos and stock data.\medskip

\noindent\textbf{Language Modeling}. Since the web neural network functions on continuous data, it can take in sequences of language tokens and perform many-to-many operations such as translation. Not only that, by embedding tokens and adding extra layers after the output, the model can generate sentences where the total number of timesteps will be the number of tokens generated.\medskip

In summary, we presented web neural network: a new versatile neural network model that can operate on continuous data while preserving context. We also provided naive and vectorized algorithms for the forward pass of the model and showed that the model can learn a process of classification that predicts an output incrementally over a certain number of timesteps. This constructs the foundational work for any future research projects on this type of network and the incremental classification phenomenon can be studied further to understand how machines classify data.

{\small
\bibliographystyle{ieee_fullname}
\bibliography{egbib}
}

\end{document}